\documentclass[10pt,twocolumn]{article}
\usepackage[letterpaper,margin=0.75in,columnsep=0.25in]{geometry}
\usepackage[T1]{fontenc}
\usepackage{mathptmx}
\usepackage{graphicx}
\usepackage{amsmath,amssymb}
\usepackage{booktabs}
\usepackage{tabularx}
\usepackage{makecell}
\usepackage[numbers,sort&compress]{natbib}
\usepackage[font=small,labelfont=bf,labelsep=period]{caption}
\usepackage{titlesec}
\usepackage[hidelinks]{hyperref}
\usepackage{xurl}
\usepackage{microtype}

\titleformat{\section}{\normalfont\large\bfseries}{\thesection.}{0.5em}{}
\titleformat{\subsection}{\normalfont\normalsize\itshape}{\thesubsection.}{0.5em}{}
\titlespacing*{\section}{0pt}{1.2ex plus .5ex}{0.6ex}
\titlespacing*{\subsection}{0pt}{1.0ex plus .4ex}{0.4ex}
\newcommand{\chat}{\hat{c}}

\begin{document}

\twocolumn[
\begin{center}
{\LARGE\bfseries Closed-loop sweep planning for robotic 3D laser inspection\\[2pt] with a generative defect model\par}
\vspace{10pt}
{\large Yue Cao, Ruo-Syuan Mei, Yujing Yang, Jeffrey A. Abell, Chenhui Shao\textsuperscript{*}\par}
\vspace{4pt}
{\normalsize\itshape Department of Mechanical Engineering, University of Michigan, Ann Arbor, MI 48109, USA\par}
\vspace{2pt}
{\small \textsuperscript{*}Corresponding author. E-mail address: \href{mailto:chshao@umich.edu}{chshao@umich.edu} (C. Shao).\par}
\end{center}
\vspace{4pt}
\begin{abstract}
\noindent Robotic laser-line inspection cells check manufactured parts against their computer-aided design (CAD) models. They usually scan the entire part along a taught path, even though the cell only has to decide whether a defect is present and whether its size is within tolerance. This paper presents a closed-loop inspection method that scans only what this decision requires, so that the inspection time per part is shortened at a prescribed probability of a wrong decision. A generative defect model completes the partially scanned part into samples of the full defect field, which act as competing hypotheses about the defect. A planner then selects each next continuous sweep according to how quickly its predicted data would separate the current conclusion from these rivals. A calibrated stopping rule ends the inspection once the conclusion is confident. In a simulated robot--turntable cell, the method was correct in at least 96\,\% of the inspections of every gear class and concluded faster than all alternatives on defective gears. On broken gears it needed 16\,s on average, compared with 22\,s for information-gain planning and 41--47\,s for industrial raster and coarse-to-fine scans. Compared with industrial fixed scans, the shorter inspection time raises the throughput of the cell and thus lowers the inspection cost per part.

\vspace{4pt}
\noindent\textit{Keywords:} Robotic inspection; Active sensing; Measurement planning; Laser line scanning; 3D metrology; Generative models; Quality control
\end{abstract}
\vspace{10pt}
]

\section{Introduction}

Robotic 3D scanning has become a standard way to inspect manufactured parts against their computer-aided design (CAD) models, but its cycle time is set by the size of the part rather than by the decision the inspection has to make. A robot, often paired with a turntable for rotational parts, moves a laser line profiler along a taught path that covers the whole part at a fixed standoff. The resulting point cloud is then compared with the nominal geometry~\cite{b1,b2}. For gears, such areal optical measurement complements tactile gear metrology with dense data on every tooth~\cite{b3}. The quality decision, however, is much narrower. The cell only has to determine whether a defect is present and whether its size lies within tolerance. Most parts are defect-free, and on the others only the neighbourhood of a defect matters. Most of a full scan therefore contributes nothing to this decision.

Existing scan planning and point-cloud analysis make the scan complete and accurate, but they do not decide which measurement the decision needs next or when the data suffice. CAD-based planners choose scanner poses and paths that cover the surface with bounded measurement uncertainty~\cite{b1,b4,b5}. In coordinate metrology, adaptive sampling selects the next probing location from a kriging or Gaussian-process model of the surface~\cite{b6,b7,b8}. On the analysis side, deep networks detect and classify gear defects from complete point clouds~\cite{b9}, supported by the synthetic MFGNet-Gear dataset of defect-free and defective gears~\cite{b10}. Because these methods target surface reconstruction or the analysis of a finished scan, none of them adapts the acquisition to the inspection decision.

Sequential hypothesis testing describes how measurements should be chosen for a decision, but two properties of laser inspection prevent its direct use. The theory selects, at each step, the experiment that best separates the leading hypothesis from its hardest rivals, and it stops once the probability of error is small enough~\cite{b11,b12}. In inspection, however, defect shapes have no parametric form, so the hypotheses cannot be enumerated in advance. In addition, a laser line profiler measures while it moves. Its actions are therefore continuous sweeps whose coverage, point density and noise depend on standoff and speed, not choices from a finite set of experiments.

This paper resolves both obstacles by coupling a generative defect model, which supplies the hypotheses, with a planner that scores continuous sweeps by how fast they separate them. The contributions are threefold. (i)~A generative defect model estimates the complete defect field from partial scans, and its samples serve as non-parametric hypotheses about the presence, type, location and size of a defect. (ii)~A planner over continuous scan-while-moving sweeps maximizes the predicted rate at which the rivals of the current conclusion lose posterior mass, computed from the CAD model and the sensor physics. A calibrated stopping rule ends the inspection. (iii)~An evaluation in a simulated robot--turntable cell on MFGNet-Gear compares the method with industrial fixed scans and with coverage and information-gain planning, and ablates each component.

The resulting loop concludes fastest on defective gears, with an observed error rate of at most 4\,\% in every class. Its advantage is largest on broken gears, whose narrow crack opening can only be measured with near and dense sweeps.

The remainder of this paper is organized as follows. Section~2 reviews related work. Section~3 formulates the problem, Section~4 presents the method, Section~5 describes the experiments, Section~6 discusses the results, and Section~7 concludes.

\section{Related work}

\subsection{Scan planning for optical inspection}

Scan planning for optical inspection selects sensor poses and paths that capture the whole part with the required data quality. Model-based planners derive scanner placements and paths from the CAD model under field-of-view, depth-of-field and incidence-angle constraints~\cite{b4,b13}, later with bounds on the measurement uncertainty~\cite{b5}, as reviewed in~\cite{b1}. For robotic line scanners, coverage path planning shortens the scan while still covering the whole surface~\cite{b14}, and point-cloud quality assessment directs rescans of sparsely sampled regions~\cite{b15}. Next-best-view planners in robotics choose each new view by expected coverage or volumetric information gain~\cite{b16,b17}, or by learned reconstruction uncertainty~\cite{b18}, and a generative object prior has been used to plan all views in one shot~\cite{b19}. Reinforcement learning has also been used to select inspection viewpoints~\cite{b20}, and deterministic evidence gates let a robotic inspection system issue a pass only after its measurement and coverage checks have succeeded~\cite{b21}. All of these methods aim at a complete reconstruction from discrete views acquired at rest, so they cannot shorten the scan to the evidence that a single accept-or-reject decision needs.

\subsection{Adaptive sampling and sequential testing}

Adaptive sampling and sequential testing choose each measurement from what has been observed so far. In coordinate metrology, kriging models place the next probing point where the predicted geometric deviation and its uncertainty are informative~\cite{b6}. Hierarchical or multi-task Gaussian-process models and dynamic sampling designs allocate measurements in manufacturing processes~\cite{b7,b8,b22}. More generally, sequential design of experiments~\cite{b11} and active sequential hypothesis testing~\cite{b12} identify the true hypothesis with a prescribed error probability $\alpha$ at the smallest expected cost. For finitely many hypotheses, this cost grows as $\log(1/\alpha)$ divided by the best achievable rate of information against the closest alternative, and efficient policies concentrate on separating the leading hypothesis from its hardest rivals~\cite{b23,b24}. Bayesian experimental design pursues related goals with posterior-based utilities~\cite{b25} and has been amortized for downstream decisions~\cite{b26}. The surrogate models of metrology describe smooth surfaces rather than local defects, and the testing frameworks assume enumerable hypotheses and discrete experiments.

\subsection{Learning-based defect inference}

Deep learning infers geometric defects from 3D data, and generative models increasingly represent the variability of manufacturing data. Point-cloud networks trained on CAD-generated gears recognize gear designs and defect classes~\cite{b9}, and MFGNet-Gear provides such data as a benchmark~\cite{b10}. In smart manufacturing, generative models synthesize process and quality data and represent the uncertainty of defect predictions~\cite{b27,b28}. Conditional variational autoencoders~\cite{b29,b30} learn a distribution over outputs given an observation, which suits them to completing a partially observed field. Existing defect models, however, analyze complete scans and do not provide the distribution over complete defects that an online inspection decision needs.

\section{Problem formulation}

\subsection{Parts, state and conclusions}

\begin{figure*}[t]
\centering
\includegraphics[width=\textwidth]{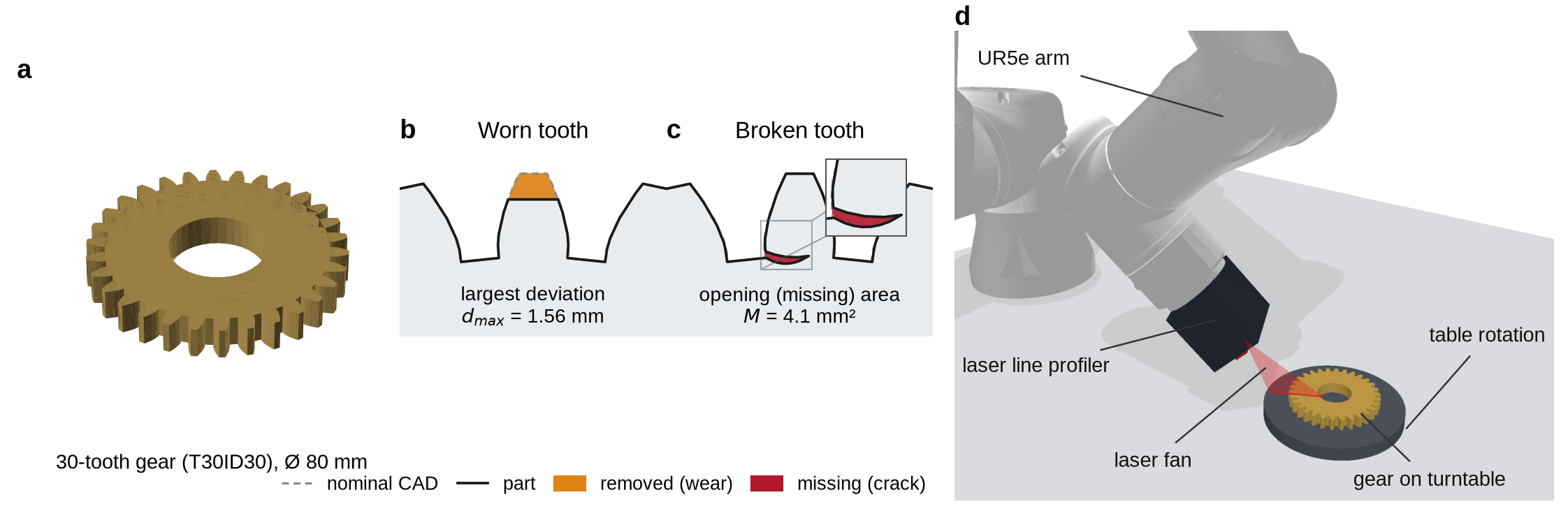}
\caption{Inspected gears and simulated cell. (a) Defect-free 30-tooth gear of the MFGNet-Gear dataset (design T30ID30, outer diameter 80\,mm), rendered from its CAD mesh. (b, c) Top-face outline of one tooth of a worn and a broken test gear (black) against the nominal outline (dashed). Shaded areas show the material removed by wear and the missing material of the crack (enlarged in the inset). The sizes to be estimated are the largest deviation $d_{\max}$ for wear and the opening area $M$ for breakage. (d) Simulated cell rendered in PyBullet, in which a UR5e arm holds a laser line profiler (red: laser fan) above the gear on a turntable that rotates the gear during a sweep.}
\label{fig:cell}
\end{figure*}

The inspection estimates a non-parametric defect field on the nominal CAD surface and reports a single conclusion about it. We consider 30-tooth gears that are defect-free, worn on one tooth, or broken by a crack that removes material from one tooth (Fig.~\ref{fig:cell}a--c). The nominal surface is divided into patches $p = 1,\ldots,P$, and the unknown state $\theta$ assigns to every patch the largest deviation $d_p$ of the part surface from the nominal surface. It also assigns the missing fraction $m_p$, the share of the patch area whose normal distance to the part surface exceeds a threshold $\tau$. A defect-free part has $\theta \approx 0$, whereas a defective part carries one local defect of unknown location and shape. A patch is anomalous when $d_p$ or $m_p$ exceeds the largest value found on defect-free training parts. The inspection reports either ``no defect'' or a defect type, its tooth and its size. For wear, the size is the largest deviation $d_{\max}$, and for breakage it is the opening area $M = \sum_p A_p m_p$ over anomalous patches, where $A_p$ is the patch area. A conclusion is correct when presence, type and tooth are right and the size is within its specification.

\subsection{Sweeps, cost and observations}

Each action is a continuous sweep that trades coverage per second against point density and noise. A six-axis arm holds the profiler at one standoff and direction while a turntable rotates the part over a chosen number of tooth pitches, from a chosen start tooth, in a chosen direction and at one of three speeds (Fig.~\ref{fig:cell}d). The duration $T(u)$ of a sweep $u$ is the scanning time plus the transfer time for the arm move, the table turn and settling. Point spacing across the laser line grows with the standoff $r$, and spacing along the sweep grows with the speed. Because fast sweeps record only a subset of the profiles and a repeated sweep records those in between, point density can be built up in steps. The range noise is $\sigma = \sigma_0 (r/r_0)^2 / \max(\cos\varphi, 0.3)$ at incidence angle $\varphi$. Near, slow sweeps therefore give dense and precise points over a short line, whereas far, fast sweeps cover more surface per second with coarser points. The task is to choose the sweeps $u_1,\ldots,u_N$, their number $N$ and a conclusion $\chat$ that minimize the expected inspection time,
\begin{equation}
\min\ \mathbb{E}\Big[\sum_{t=1}^{N} T(u_t)\Big] \quad \text{s.t.} \quad \Pr(\chat\ \text{wrong}) \le \alpha,
\label{eq:problem}
\end{equation}
where $\alpha$ is the admissible probability of a wrong conclusion.

\section{Method}

\subsection{Inspection loop and design principle}

\begin{figure*}[t]
\centering
\includegraphics[width=\textwidth]{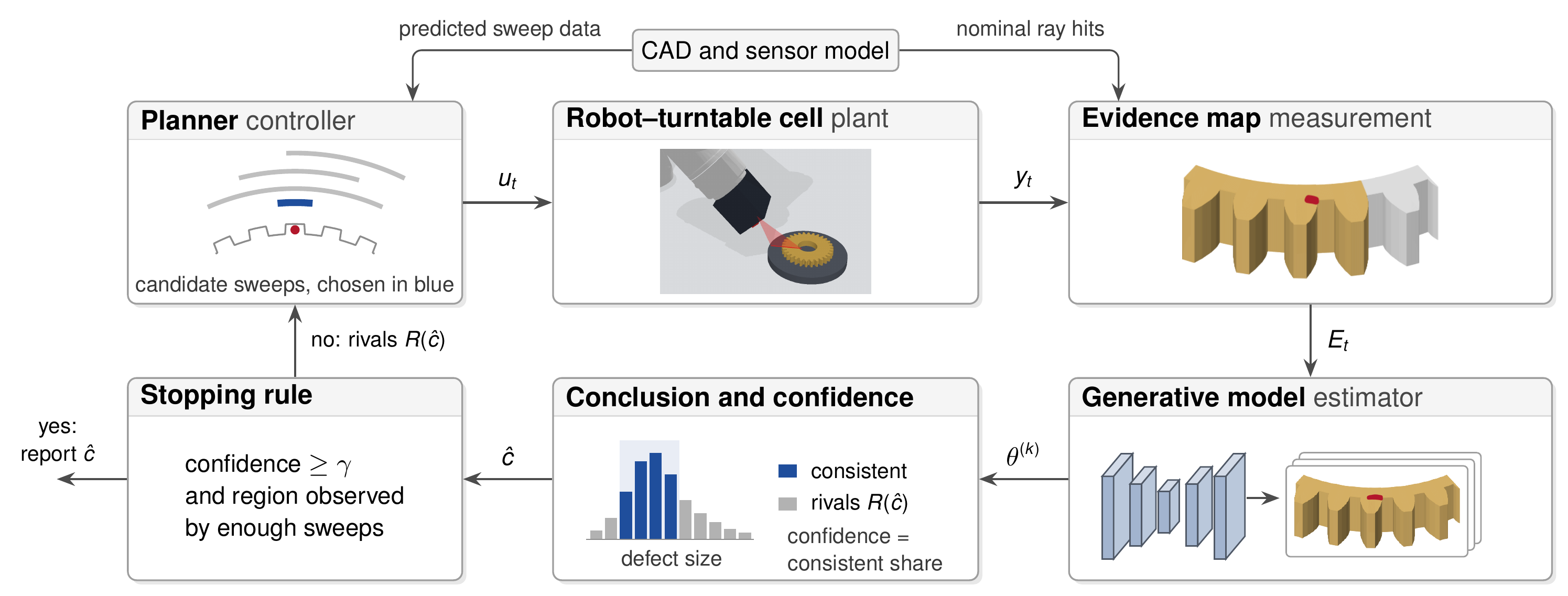}
\caption{Inspection as a closed loop. The planner chooses the next sweep $u_t$ from candidate sweeps over the tooth ring, predicting their data with the CAD and sensor model. The robot--turntable cell executes the sweep while the profiler records profiles $y_t$, and comparing each ray with its nominal hit updates the evidence map $E_t$ (gold: observed, red: missing material, gray: not yet observed). The generative model returns samples $\theta^{(k)}$ of the complete defect field, whose size distribution gives the conclusion $\chat$, its confidence (the share of consistent samples) and the rival samples $R(\chat)$. The stopping rule reports $\chat$ or passes the rivals back to the planner.}
\label{fig:loop}
\end{figure*}

The method closes a feedback loop around the cell, with the generative defect model as the estimator and the sweep planner as the controller (Fig.~\ref{fig:loop}). In each cycle, the cell executes the sweep chosen by the planner. Comparing every measured ray with the CAD model then updates an evidence map. From this map, the generative model draws samples of the complete defect field, which define the current conclusion, its confidence and its rivals. The stopping rule either reports the conclusion or passes the rivals back to the planner. Using the CAD model and the sensor model, the planner then predicts how candidate sweeps would separate these rivals.

The planner follows the principle of sequential hypothesis testing that each measurement should separate the current conclusion from its hardest rivals. For finitely many hypotheses, the smallest expected cost achievable at error probability $\alpha$ grows as $\log(1/\alpha)$ divided by the largest achievable rate, per unit cost, of Kullback--Leibler (KL) divergence between the true hypothesis and its closest alternative~\cite{b11,b12,b23,b24}. The planner therefore scores a sweep by how fast it separates the conclusion from its rivals, not by the new area it covers. This criterion adapts the scan to the remaining uncertainty. When the rivals place the defect on other teeth, a wide sweep over many teeth separates them. When they differ only in the size of a small crack, near and dense sweeps over that tooth do. The generative model is what makes the principle applicable to non-parametric defects, because its samples supply the hypotheses that the theory assumes to be enumerable.

\subsection{Evidence map}

The evidence map summarizes, for every patch, how the measured rays compare with the CAD model. Ray casting on the CAD mesh gives the nominal hit of every profiler ray. A ray counts as missing when it has no hit or when its measured range deviates from the nominal one by more than $\max(5\sigma, \tau/\cos\varphi)$. For each patch, the map accumulates point counts, residuals, the largest measured deviation, the noise level, the visible area, the number of observing sweeps and the missing fraction. The missing fraction is stored both pooled over all sweeps and as its largest single-sweep value. Sweeps whose direction cannot see into a defect therefore do not dilute earlier evidence of missing material.

\subsection{Generative defect model}

\begin{figure*}[t]
\centering
\includegraphics[width=\textwidth]{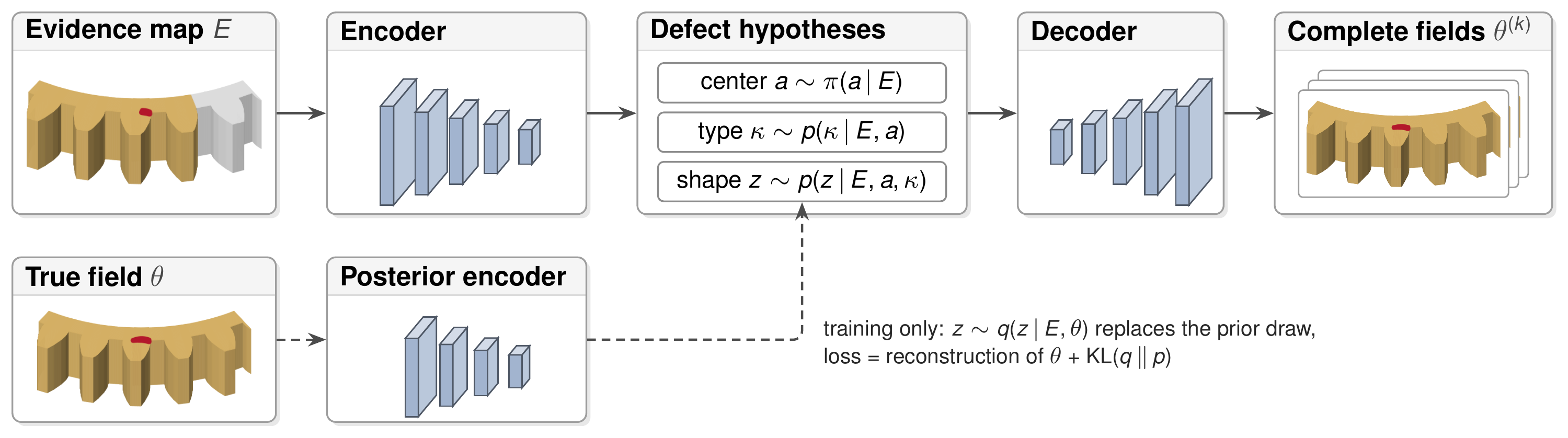}
\caption{Structure and training of the generative defect model. The encoder reads the evidence map $E$ of a partial scan (gold: observed, red: missing material, gray: not yet observed). Three heads sample the defect center $a$, the type $\kappa$ and the latent shape $z$, and the decoder maps each draw to a complete defect field $\theta^{(k)}$. Dashed paths are used only in training, where a posterior encoder that also sees the true field $\theta$ supplies $z$, and the loss combines the reconstruction of $\theta$ with the KL divergence between the posterior and the prior of $z$. The gear sector is rendered from the CAD mesh of a broken test gear.}
\label{fig:model}
\end{figure*}

The generative model completes the defect field from partial evidence by sampling, in turn, where the defect is, what type it is and what shape it has (Fig.~\ref{fig:model}). It factorizes the posterior over complete defect fields as
\begin{equation}
\begin{aligned}
p(\theta \mid E) ={}& \pi(\text{none} \mid E) \\ &\times \int p(z \mid E, \text{none})\, p(\theta \mid E, \text{none}, z)\, dz \\
&+ \sum_{a \neq \text{none},\, \kappa} \pi(a \mid E)\, p(\kappa \mid E, a) \\
&\times \int p(z \mid E, a, \kappa)\, p(\theta \mid E, a, \kappa, z)\, dz,
\end{aligned}
\label{eq:posterior}
\end{equation}
where $a$ is the defect-center patch (or ``none''), $\kappa$ the defect type and $z$ a latent shape variable. A graph neural network encodes the evidence map together with the patch geometry. It does so through residual message-passing layers~\cite{b31} on the nearest-neighbor graph of the patches, and its messages depend only on the relative cylindrical coordinates of neighbouring patches. The model is therefore equivariant to rotations of the part by whole teeth. Three heads give $\pi(a \mid E)$, $p(\kappa \mid E, a)$ and a Gaussian prior $p(z \mid E, a, \kappa)$. A decoder shared by all patches then maps the patch features, the position relative to $a$, the type and $z$ to $d_p$ and $m_p$. Each draw of $(a, \kappa, z)$ thus yields one complete defect field, including the regions not yet scanned.

The model is trained as a conditional variational autoencoder on simulated partial scans of training parts. During training, a posterior encoder $q(z \mid E, \theta)$ that also sees the true field supplies $z$~\cite{b29,b30}. The loss combines cross-entropy terms for $a$ and $\kappa$, an L1 loss on $d_p$, a binary cross-entropy on $m_p$ and the KL divergence between $q$ and the prior. Flooring the KL term per latent dimension (free bits~\cite{b32}) keeps the latent variable in use, so that the samples retain the spread of defect shapes. The model is first trained on evidence maps after random sweep sequences. The planner, however, concentrates its sweeps on the defect, so the states it visits differ from random ones. The model is therefore fine-tuned on states visited by the planner on training parts, in the spirit of dataset aggregation (DAgger)~\cite{b33}. Finally, the spread of the prior over $z$ is scaled by a temperature chosen on validation parts.

\subsection{Conclusion, confidence and rivals}

The posterior samples define the conclusion, its confidence and the rivals that the planner must eliminate. At each step, the posterior is represented by $K$ equally weighted samples. Each sample implies its own conclusion, either ``none'' or the tooth of its defect center, its type and its size. The conclusion $\chat$ is the most frequent tooth and type, or ``none'', with the median size of its samples. A sample is consistent with $\chat$ if it has the same type, lies within one pitch of the tooth of $\chat$ and has a size within tolerance of $\chat$. For $\chat = \text{``none''}$, a sample is consistent if it is defect-free. The confidence is the share of consistent samples, and the remaining samples form the rival set $R(\chat)$.

\subsection{Sweep planning}

The planner scores each candidate sweep by the predicted rate at which its data reduce the posterior mass of the rivals. For a candidate sweep, ray casting on the CAD model gives the expected number of points $n_p$ and the noise $\sigma_p$ on each patch. Treating points as independent, each point is missing with the probability given by the missing fraction of a sample, and points on deviating material carry Gaussian range noise $\sigma_p$. The expected log-likelihood ratio of the data of the sweep between a consistent sample $i$ and a rival $j$ then decomposes over patches and is approximated by
\begin{equation}
\begin{aligned}
\mathrm{KL}_{ij}(u) &= \sum_p \min\{ n_p \ell_{ij,p},\, C \}, \\
\ell_{ij,p} &= w_p \frac{(d_{i,p} - d_{j,p})^2}{2\sigma_p^2} + \mathrm{KL}_{\mathrm{Bern}}(m_{i,p} \,\|\, m_{j,p}),
\end{aligned}
\label{eq:kl}
\end{equation}
where the sum runs over the patches on which the samples vary. The weight $w_p = (m_{i,p} + m_{j,p})/2$ is the mean share of points that deviate under the two samples and therefore carry depth information. $\mathrm{KL}_{\mathrm{Bern}}$ is the KL divergence between the Bernoulli distributions of a point being missing, and the cap $C$ keeps a single patch from dominating. After the sweep, the log posterior weight of rival $j$ is expected to decrease by $\mathrm{KL}_j$, the mean of $\mathrm{KL}_{ij}$ over the consistent samples. Its predicted remaining weight is therefore $\exp(-\mathrm{KL}_j)$. The planner executes the candidate that maximizes the predicted rate of rival-mass reduction per second,
\begin{equation}
J(u) = -\log\Big( \operatorname*{mean}_{j \in R(\chat)} e^{-\mathrm{KL}_j(u)} \Big) \Big/ T(u).
\label{eq:score}
\end{equation}

The continuous sweep space is searched by sampling candidates at every decision, so no phases, standoffs or speeds are prescribed. The candidates comprise random sweeps over all standoff--direction families, start teeth, directions, lengths and speeds, continuations of the current sweep, and short sweeps around the tooth of $\chat$. When no rivals remain but the inspection may not stop yet, the planner takes the candidate that observes the most patches per second. After each executed sweep, the evidence map is updated with the actual measurements and the posterior is sampled anew. The predicted rival-mass reduction therefore serves only to choose the sweep.

\subsection{Stopping rule}

The inspection stops once the conclusion is confident and its region has been observed by enough sweeps. The confidence condition controls the error rate. It requires the confidence to reach a threshold $\gamma$ for the type of the conclusion (none, wear or breakage). Each threshold is the smallest value for which validation inspections stopped with that type have an error rate of at most $\alpha$. The evidence condition guards against confident conclusions from few views. It requires the tooth of the conclusion and both neighbors to have been observed by a minimum number of sweeps, and for ``no defect'' it requires every tooth to have been observed at least once. An inspection that reaches the time budget ends with its current conclusion, so every inspection returns a decision.

\section{Experimental setup}

\subsection{Data and splits}

The experiments use the 30-tooth design of MFGNet-Gear with defect-free, worn and broken parts. The dataset~\cite{b10} provides CAD meshes of 99 defect-free, 100 worn and 100 broken gears of design T30ID30. Within each class, the first 60\,\% of the parts train the generative model, the next 15\,\% serve as validation parts for the temperature and the stopping thresholds, and the last 25\,\% are test parts. Each validation and test part is inspected from two random table start angles, which gives 30 validation and 50 test inspections per class. The specifications are $\pm 0.1$\,mm on $d_{\max}$ for wear and $\pm 20$\,\% on $M$ for breakage. The worn test teeth deviate by up to 2.4\,mm, whereas the test cracks are sub-millimeter wide with opening areas of 3.4--4.4\,mm$^2$.

\subsection{Simulated cell and sensor}

The simulated cell models a UR5e arm, a turntable and an idealized laser line profiler. The arm is simulated in PyBullet~\cite{b34}. The profiler projects a 16$^\circ$ fan of 320 rays, records 100 profiles per second over a range of 120--220\,mm and has $\sigma_0 = 0.02$\,mm at $r_0 = 170$\,mm. Reflections and registration errors are not modeled. Sweeps use 40 standoff--direction families, formed by five standoffs (130, 145, 160, 180 and 200\,mm) and eight directions, and three speeds, the slowest taking 0.48\,s per pitch. The arm moves at $\pi$\,rad\,s$^{-1}$ per joint and the table at 120$^\circ$\,s$^{-1}$. Each transfer adds 0.2\,s of settling, and the time budget is 75\,s. The missing threshold $\tau$ is 0.1\,mm.

\subsection{Implementation}

The generative model is small enough to be trained in about half an hour and queried while a sweep is running. The encoder has six residual message-passing layers with 64 channels on the eight-nearest-neighbor graph of the patches. The posterior encoder has three such layers, and the latent variable has 16 dimensions with a free-bits floor of 0.1\,nat per dimension. Training takes 6,000 steps with batches of 32 states sampled on the fly (AdamW, learning rate $10^{-3}$ with a one-cycle schedule, weight decay $10^{-4}$), about 20\,min on one RTX 5070 Ti graphics processing unit. Fine-tuning takes 3,000 steps at $3 \times 10^{-4}$, with half of each batch drawn from planner trajectories on the training parts, about 9\,min. The planner uses $K = 64$ samples, a cap $C = 20$, and 300 random candidates plus the continuations and short sweeps at each decision, each candidate lasting at least 0.5\,s. A planning decision takes about 0.13\,s, less than the shortest sweep, so it can run while the previous sweep is executed. The stopping thresholds are chosen on a grid from 0.50 to 1.00 with $\alpha = 0.05$ and a minimum of four sweeps over the region of the conclusion.

\subsection{Baselines and ablations}

The baselines comprise two industrial fixed scans and three planners that share our generative model and stopping rule. The raster scan runs slow full rings at the nearest standoff in the families that together observe all defect-prone patches. The coarse-to-fine scan runs fast full rings at the farthest standoff and then rescans, with slow short sweeps at the nearest standoff, the teeth whose evidence exceeds the largest value found on defect-free training parts. Both fixed scans report the conclusion of our generative model once their plan is complete. Among the planners, coverage planning maximizes the new visible area per second, and information-gain planning maximizes the entropy reduction of the per-patch anomaly probability per second, after~\cite{b16}. A hand-designed two-phase rule covers the part until a patch is likely anomalous and then maximizes the points per noise variance on such patches per second. Each planner uses its own calibrated thresholds. The ablations restrict our planner to one standoff or to stop-and-go single-pitch sweeps, vary the minimum number of sweeps in the stopping rule, and replace the generative model by one with independent patches. The time to a conclusion is the stopping time for planners and the plan duration for fixed scans. Error bounds are one-sided 95\,\% Clopper--Pearson bounds~\cite{b35}.

\begin{figure*}[t]
\centering
\includegraphics[width=\textwidth]{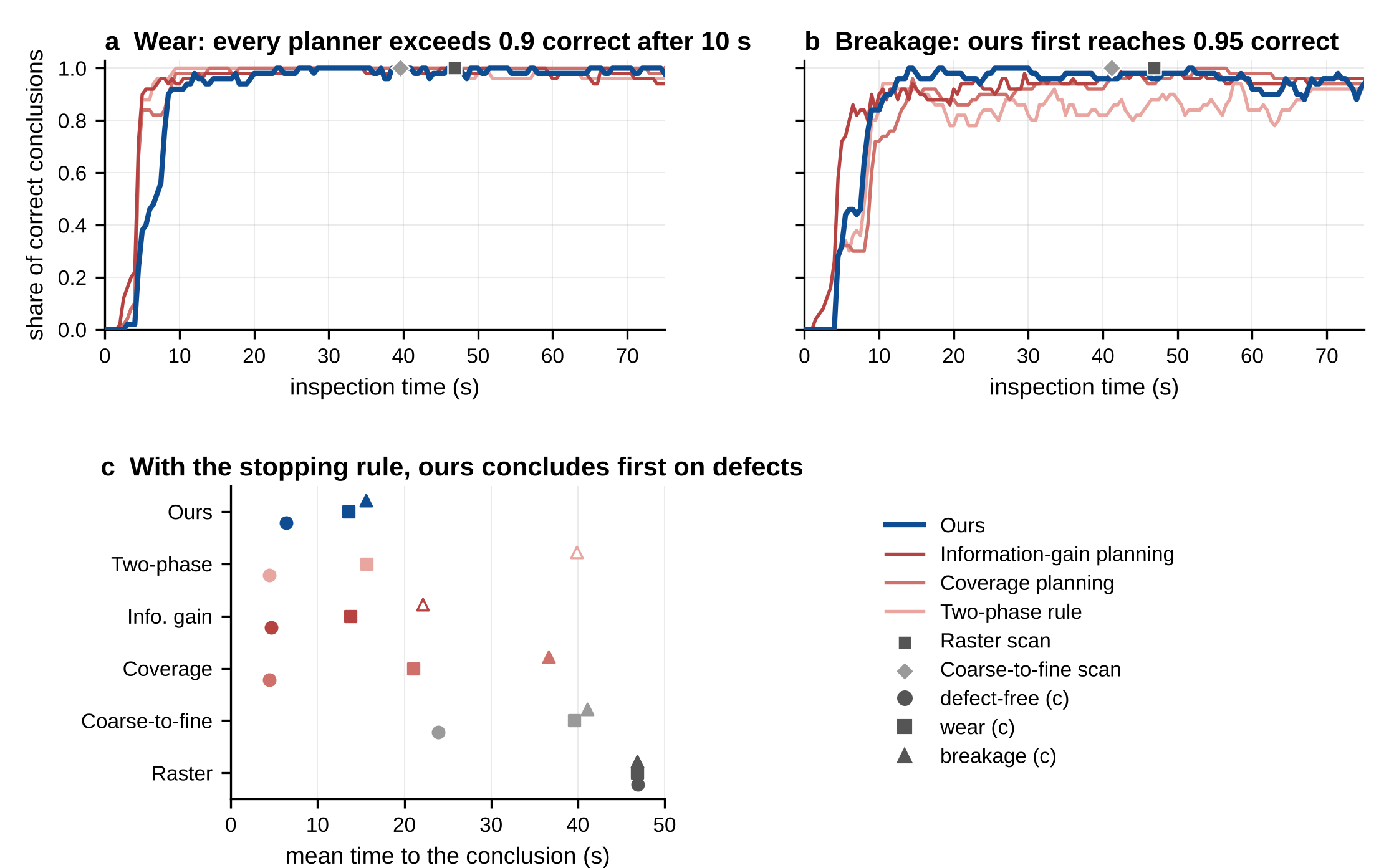}
\caption{Accuracy over time and time to the conclusion on the 50 test inspections per class. (a, b) Share of correct conclusions (presence, type, tooth and size within specification) versus inspection time for worn and broken gears, evaluated for the conclusion each planner would report at that time. Markers show the fixed scans at their plan duration. (c) Mean time to the conclusion with the calibrated stopping rule, per class. Open markers mark fewer than 48 correct conclusions out of 50.}
\label{fig:time}
\end{figure*}

\section{Results and discussion}

\subsection{Time and accuracy}

Our planner concluded fastest on defective gears, with an observed error rate of at most 4\,\% in every class (Table~\ref{tab:main}). The observed error rates were 0, 4 and 2\,\% on defect-free, worn and broken gears, and their one-sided 95\,\% upper bounds, from 50 inspections of 25 test parts per class, were 6--12\,\%. The fixed scans were always correct but took 24--47\,s. The gain is largest on broken gears, where our planner concluded after 15.6\,s, whereas the next fastest planner, information-gain planning, took 22.1\,s and made three errors. On worn and defect-free gears the adaptive planners were close. Ours was about 2\,s slower on defect-free gears because it often ran the first ring at medium speed.

\begin{table*}[t]
\centering
\caption{Mean $\pm$ standard deviation of the time to the conclusion (s) and number of correct conclusions out of 50 test inspections per class with the calibrated stopping rule. Fixed scans end with their plan.}
\label{tab:main}
\begin{tabular}{lccc}
\toprule
Method & Defect-free & Wear ($\pm$0.1\,mm) & Breakage ($\pm$20\,\%) \\
\midrule
Raster scan (fixed plan) & 46.9 $\pm$ 0.3 / 50 & 46.8 $\pm$ 0.2 / 50 & 46.9 $\pm$ 0.2 / 50 \\
Coarse-to-fine scan (fixed plan) & 23.9 $\pm$ 19.8 / 50 & 39.6 $\pm$ 12.2 / 50 & 41.1 $\pm$ 12.6 / 50 \\
Coverage planning & 4.4 $\pm$ 0.3 / 50 & 21.0 $\pm$ 7.0 / 49 & 36.6 $\pm$ 20.1 / 49 \\
Information-gain planning & 4.6 $\pm$ 1.1 / 50 & 13.8 $\pm$ 2.2 / 49 & 22.1 $\pm$ 15.7 / 47 \\
Two-phase rule & 4.4 $\pm$ 0.3 / 50 & 15.7 $\pm$ 2.0 / 49 & 39.8 $\pm$ 25.6 / 46 \\
\textbf{Ours} & \textbf{6.4 $\pm$ 1.9 / 50} & \textbf{13.5 $\pm$ 3.7 / 48} & \textbf{15.6 $\pm$ 6.4 / 49} \\
\bottomrule
\end{tabular}
\end{table*}

Breakage separates the planners, because the crack opening must still be measured to $\pm 20$\,\% once the crack has been found (Fig.~\ref{fig:time}). On worn gears every planner exceeded 0.9 correct after 10\,s. On broken gears, by contrast, our planner was the first to reach 0.95 correct, at 12.5\,s, and it stayed at or above 0.92 between 12 and 35\,s. In the same interval, information-gain planning fell to 0.86, the two-phase rule to 0.78 and coverage planning to 0.76. Only our planner scores sweeps by how well they separate samples that differ in opening area, so it keeps measuring the crack until its size is settled. Under a stricter $\pm 10$\,\% specification on the opening area, every method stayed at or below 0.80 correct. This specification therefore lies near the measurement capability of the simulated sensor.

\subsection{Posterior during an inspection}

\begin{figure*}[t]
\centering
\includegraphics[width=\textwidth]{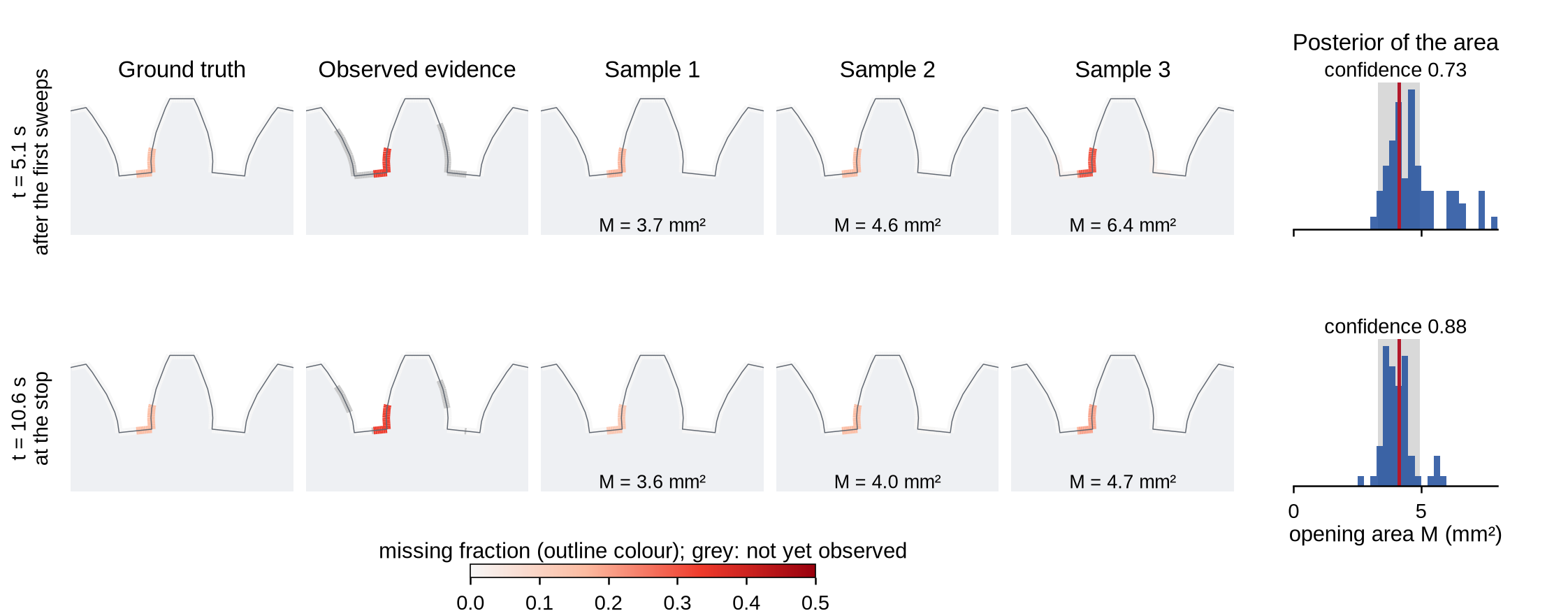}
\caption{Generative posterior during a test inspection of a broken gear, after two sweeps (5.1\,s, top) and at the stop (10.6\,s, bottom). Each view shows the outline of the defective tooth and its neighbors, coloured on the side faces by the missing fraction, for the ground truth, the observed evidence (largest single-sweep missing fraction, gray: not yet observed) and three of the 64 samples with low, median and high opening area $M$. Right: distribution of $M$ over the samples (red line: true area, gray band: $\pm 20$\,\% specification).}
\label{fig:posterior}
\end{figure*}

The generative posterior completes the crack where it has not been observed and narrows its size distribution as near sweeps are added (Fig.~\ref{fig:posterior}). After the first two sweeps of the inspection shown, the samples already place the crack on the correct tooth, but their opening areas still range from about 3 to 8\,mm$^2$. The rivals are then mostly samples of the same crack with too large an area, so the planner turns to near sweeps over that tooth. Five further sweeps narrow the distribution around the true area, and the inspection stops at 10.6\,s.

\begin{figure*}[t]
\centering
\includegraphics[width=\textwidth]{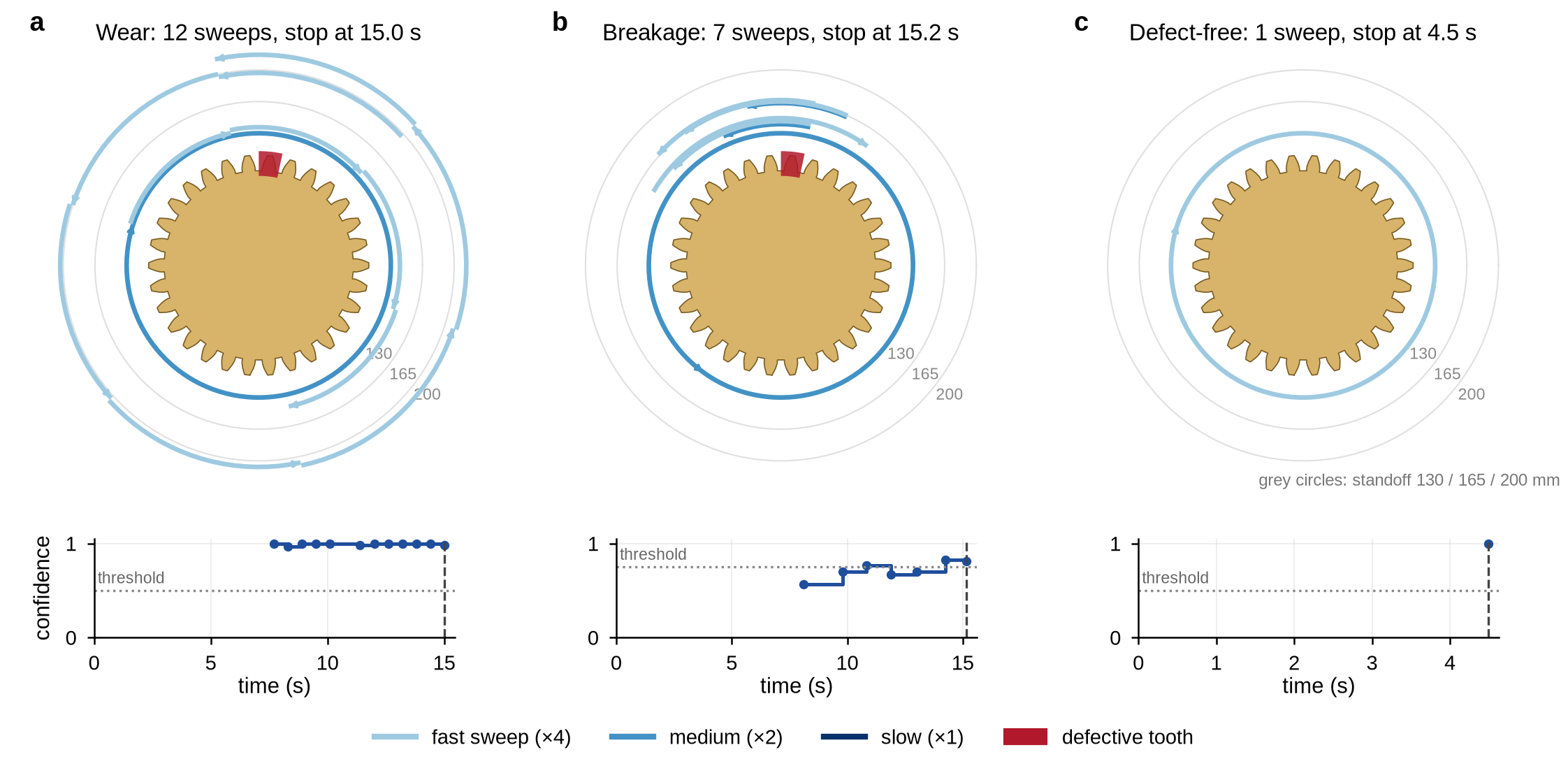}
\caption{Sweep sequences of our planner up to the stop, one test inspection per class, viewed from above. Each arc is one sweep (arrow: direction), its radius grows with the standoff (gray circles: standoff scale from 130 to 200\,mm), and successive sweeps are offset slightly so that repeated sweeps remain visible. Color shows the speed, and red marks the defective tooth. Below: confidence after each sweep, the stop threshold $\gamma$ of the type of the conclusion (dotted) and the stop (dashed). Because the stop also requires four sweeps over the defect region, it can come later than the first threshold crossing, as on the worn gear (a).}
\label{fig:sweeps}
\end{figure*}

\subsection{Where the planner looks}

Without being told to, the planner chooses the standoff by defect type, in agreement with the sensor physics (Fig.~\ref{fig:sweeps}, Table~\ref{tab:behavior}). The first sweep is usually a full ring, which suffices for defect-free gears. On defective gears the planner then returns to the defect with short sweeps. Of these, 28\,\% are farther than 160\,mm on worn gears but only 3\,\% on broken gears. Worn teeth deviate by millimeters, which a far sweep measures with a longer laser line and a noise well below the $\pm 0.1$\,mm tolerance. The opening of a sub-millimeter crack, in contrast, needs the dense points of near sweeps. On defective gears, most sweep time is spent at the fastest speed, because repeated fast passes build up point density in steps and let the inspection stop as soon as it suffices. All wrong conclusions were sizes just outside the specification, with no missed defects, false alarms or wrong teeth.

\begin{table}[t]
\centering
\caption{Our planner at the stop (50 test inspections per class). Size error: $|\hat{d}_{\max} - d_{\max}|$ for wear and $|\hat{M} - M|/M$ for breakage. Speed shares refer to sweep time, and transfers include arm moves, table turns and settling.}
\label{tab:behavior}
\footnotesize
\setlength{\tabcolsep}{3pt}
\begin{tabularx}{\columnwidth}{@{}>{\raggedright\arraybackslash}Xccc@{}}
\toprule
 & Defect-free & Wear & Breakage \\
\midrule
Missed defects, false alarms, wrong type or tooth & 0 & 0 & 0 \\
Size outside the specification & -- & 2 & 1 \\
Size error, median (90th percentile) & -- & \makecell{0.036 (0.066)\\mm} & \makecell{5.6 (14.1)\\\%} \\
Sweeps per inspection (mean) & 1.1 & 8.7 & 9.0 \\
First sweep is a full ring & 94\,\% & 88\,\% & 90\,\% \\
Later sweeps farther than 160\,mm & -- & 28\,\% & 3\,\% \\
Sweep time at the fastest speed & 32\,\% & 74\,\% & 58\,\% \\
Transfers, share of inspection time & 24\,\% & 23\,\% & 29\,\% \\
\bottomrule
\end{tabularx}
\end{table}

\subsection{Ablations}

The generative model is what makes size estimation from partial scans possible, and continuous sweeps with a free standoff are what make the inspection fast (Table~\ref{tab:ablation}). On partial scans with the defect in view, a model with independent patches put only 0.14 (wear) and 0.54 (breakage) of the size estimates within specification, against 0.98 and 0.92 for the generative model. Stop-and-go sweeps slowed every class and lowered the accuracy on defects. A single standoff kept the conclusions accurate but slowed breakage to 54\,s and prevented every worn-gear inspection from stopping before the budget. The evidence condition of the stopping rule is also necessary. Without it, 10 of the 50 worn-gear inspections ended with a wrong conclusion, because the confidence from the first views of a worn tooth was not reliable on the test parts. Four sweeps over the region met the target in every class, whereas five only slowed the defect classes.

\begin{table}[t]
\centering
\caption{Ablations of the planner and the stopping rule: mean time to the conclusion and number of correct conclusions out of 50 test inspections. Stopping thresholds are recalibrated on the validation parts for each variant, and mean times above the 75\,s budget arise because a sweep in progress is completed.}
\label{tab:ablation}
\footnotesize
\setlength{\tabcolsep}{3pt}
\begin{tabularx}{\columnwidth}{@{}>{\raggedright\arraybackslash}Xccc@{}}
\toprule
Variant & Defect-free & Wear & Breakage \\
\midrule
\textbf{Ours (four sweeps over the region)} & \textbf{6.4\,s / 50} & \textbf{13.5\,s / 48} & \textbf{15.6\,s / 49} \\
Single standoff (160\,mm) & 6.5\,s / 50 & 76.4\,s / 50 & 54.2\,s / 49 \\
Stop-and-go single-pitch sweeps & 60.3\,s / 50 & 36.8\,s / 46 & 75.3\,s / 42 \\
No minimum number of sweeps & 6.4\,s / 50 & 6.2\,s / 40 & 15.2\,s / 49 \\
Three sweeps over the region & 6.4\,s / 50 & 10.8\,s / 47 & 12.5\,s / 49 \\
Five sweeps over the region & 6.4\,s / 50 & 16.4\,s / 48 & 20.1\,s / 49 \\
\bottomrule
\end{tabularx}
\end{table}

Fine-tuning on the states that the planner visits removes a drift of the size estimate along the inspection. A model trained only on random sweep sequences overestimated the opening area early in the validation inspections and underestimated it late. The error was $+8$\,\% after 10\,s and $-9$\,\% after 75\,s, because the planner concentrates its sweeps on the defect and so visits states unlike random ones. Fine-tuning on planner states, together with the largest single-sweep missing fraction as an evidence feature, reduced the drift to $+5$\,\% and 0\,\%.

\section{Conclusions}

Robotic laser inspection can be shortened to what the quality decision requires when a generative defect model and a continuous sweep planner are coupled in a closed loop. The model turns partial evidence into samples of complete defect fields, and the planner scores sweeps by how fast they separate the current conclusion from its rivals. Together they let the loop decide where to scan, from how far, how fast and when to stop. On defective gears the loop concluded faster than industrial fixed scans and alternative planners while keeping the observed error rate below the prescribed level in every class. Defect-free parts were mostly accepted after a single sweep, so the scanning effort went to the parts that carry a defect. Rival separation alone led the planner to near sweeps for cracks and far sweeps for wear, without prescribed phases. The generative completion proved essential, since the size of a partially observed defect could not be estimated with independent patches. Continuous sweeping and the free choice of standoff were equally necessary, since stop-and-go sweeps or a single standoff multiplied the inspection time. The loop needs only a CAD model, a sensor model and simulated training scans of the part family, so it fits cells that inspect families of similar parts against their CAD models.

Future work will extend the method beyond the conditions for which it was developed. The generative model proposes only the defect kinds it was trained on, so an unseen defect type, multiple defects or a new part design could lead to a confident wrong conclusion. Training it on broader families of defects and parts will address this risk. The idealized sensor model raises a similar issue, because it omits reflections and registration errors that could also be read as defects. The next step is therefore to run the loop in a physical robot--turntable cell with a measured sensor model. The generative model can then be fine-tuned on the scan states of this cell and the stopping thresholds, which control the error rate empirically rather than with a guarantee, calibrated on real parts.


\begin{thebibliography}{99}
\bibitem{b1} Peuzin-Jubert M, Polette A, Nozais D, Mari JL, Pernot JP. Survey on the view planning problem for reverse engineering and automated control applications. Comput-Aided Des 2021;141:103094.
\bibitem{b2} Chen Z, Zhang Y, Jalil Piran F, Zhou Q, Tang J, Imani F. ScanBot: A benchmark for precision robotic surface scanning with industrial laser profilers. arXiv:2505.17295; 2025.
\bibitem{b3} Goch G. Gear metrology. CIRP Ann 2003;52(2):659--95.
\bibitem{b4} ElMaraghy HA, Yang X. Computer-aided planning of laser scanning of complex geometries. CIRP Ann 2003;52(1):411--4.
\bibitem{b5} Mahmud M, Joannic D, Roy M, Isheil A, Fontaine JF. 3D part inspection path planning of a laser scanner with control on the uncertainty. Comput-Aided Des 2011;43(4):345--55.
\bibitem{b6} Ascione R, Moroni G, Petr\`o S, Romano D. Adaptive inspection in coordinate metrology based on kriging models. Precis Eng 2013;37(1):44--60.
\bibitem{b7} Yang Y, Zhang Y, Cai YD, Lu Q, Koric S, Shao C. Hierarchical measurement strategy for cost-effective interpolation of spatiotemporal data in manufacturing. J Manuf Syst 2019;53:159--68.
\bibitem{b8} Mehta M, Shao C. Adaptive sampling design for multi-task learning of Gaussian processes in manufacturing. J Manuf Syst 2021;61:326--37.
\bibitem{b9} Mei RS, Conway C, Bimrose M, King W, Shao C. Deep learning of 3D point clouds for detecting geometric defects in gears. Manuf Lett 2024;41:1324--33.
\bibitem{b10} Mei RS, Shao C. Descriptor: A synthetic 3-D gear dataset for manufacturing quality inspection (MFGNet-Gear). IEEE Data Descr 2026;3:813--21. \url{https://doi.org/10.1109/IEEEDATA.2026.3749527}.
\bibitem{b11} Chernoff H. Sequential design of experiments. Ann Math Stat 1959;30:755--70.
\bibitem{b12} Naghshvar M, Javidi T. Active sequential hypothesis testing. Ann Stat 2013;41:2703--38.
\bibitem{b13} Scott WR, Roth G, Rivest JF. View planning for automated three-dimensional object reconstruction and inspection. ACM Comput Surv 2003;35:64--96.
\bibitem{b14} Chen H, Huo S, Muddassir M, Lee HY, Liu Y, Li J, et al. PSO-based optimal coverage path planning for surface defect inspection of 3C components with a robotic line scanner. IEEE Trans Instrum Meas 2025;74:7505012.
\bibitem{b15} Li L, Zhang X. A robust assessment method of point cloud quality for enhancing 3D robotic scanning. Robot Comput-Integr Manuf 2025;92:102863.
\bibitem{b16} Isler S, Sabzevari R, Delmerico J, Scaramuzza D. An information gain formulation for active volumetric 3D reconstruction. In: Proc IEEE Int Conf Robot Autom (ICRA); 2016. p. 3477--84.
\bibitem{b17} Li L, Wan L, Zhang X. Boundary exploration of next best view policy in 3D robotic scanning. arXiv:2412.10444; 2024.
\bibitem{b18} Shen C, Zheng S, Wu B, Feng Y, Fei Y, Mei M, Jiang H, Huang X. Auto3R: Automated 3D reconstruction and scanning via data-driven uncertainty quantification. arXiv:2512.04528; 2025.
\bibitem{b19} Pan S, Jin L, Huang X, Stachniss C, Popovi\'c M, Bennewitz M. DM-OSVP++: One-shot view planning using 3D diffusion models for active RGB-based object reconstruction. arXiv:2504.11674; 2025.
\bibitem{b20} Tang W, Jahanshahi MR. Autonomous robotic inspection based on active vision and deep reinforcement learning. In: Proc Int Workshop Struct Health Monit; 2023.
\bibitem{b21} Chen Z, Ge J, Chen R, Castanier MP, Gorsich D, Imani F. Task-specified active metrological inspection with measurement-steered VLA manipulation and deterministic evidence gating. arXiv:2609.14219; 2026.
\bibitem{b22} Shao C, Jin J, Hu SJ. Dynamic sampling design for characterizing spatiotemporal processes in manufacturing. J Manuf Sci Eng 2017;139(10):101002.
\bibitem{b23} Garivier A, Kaufmann E. Optimal best arm identification with fixed confidence. In: Proc Conf Learn Theory (COLT), PMLR 49; 2016. p. 998--1027.
\bibitem{b24} Degenne R, Koolen WM. Pure exploration with multiple correct answers. In: Adv Neural Inf Process Syst 32; 2019. p. 14591--600.
\bibitem{b25} Rainforth T, Foster A, Ivanova DR, Bickford Smith F. Modern Bayesian experimental design. Stat Sci 2024;39:100--14.
\bibitem{b26} Huang D, Guo Y, Acerbi L, Kaski S. Amortized Bayesian experimental design for decision-making. In: Adv Neural Inf Process Syst 37; 2024.
\bibitem{b27} Ahsan M, Islam M, Bappy M. Generative modeling in smart manufacturing: Applications, challenges, and future directions. Manuf Lett 2025;44:1285--95.
\bibitem{b28} Mei RS, Li G, Jia S, Huang L, Shih LW, Arinez JF, et al. Synthetic data generation in smart manufacturing applications: A systematic review and outlook. J Manuf Syst 2026;89:1057--76.
\bibitem{b29} Kingma DP, Welling M. Auto-encoding variational Bayes. In: Proc Int Conf Learn Represent (ICLR); 2014.
\bibitem{b30} Sohn K, Lee H, Yan X. Learning structured output representation using deep conditional generative models. In: Adv Neural Inf Process Syst 28; 2015. p. 3483--91.
\bibitem{b31} Gilmer J, Schoenholz SS, Riley PF, Vinyals O, Dahl GE. Neural message passing for quantum chemistry. In: Proc 34th Int Conf Mach Learn (ICML), PMLR 70; 2017. p. 1263--72.
\bibitem{b32} Kingma DP, Salimans T, Jozefowicz R, Chen X, Sutskever I, Welling M. Improved variational inference with inverse autoregressive flow. In: Adv Neural Inf Process Syst 29; 2016. p. 4743--51.
\bibitem{b33} Ross S, Gordon G, Bagnell D. A reduction of imitation learning and structured prediction to no-regret online learning. In: Proc 14th Int Conf Artif Intell Stat (AISTATS), PMLR 15; 2011. p. 627--35.
\bibitem{b34} Coumans E, Bai Y. PyBullet, a Python module for physics simulation for games, robotics and machine learning. \url{http://pybullet.org}; 2016--2021.
\bibitem{b35} Clopper CJ, Pearson ES. The use of confidence or fiducial limits illustrated in the case of the binomial. Biometrika 1934;26:404--13.
\end{thebibliography}
\end{document}